\documentclass[11pt]{article}
\usepackage{array}
\usepackage{multirow}
\usepackage{hhline}
\usepackage{booktabs, makecell}
\usepackage{smile}
\usepackage[numbers]{natbib}
\usepackage[colorlinks,
           linkcolor=blue,
           anchorcolor=blue,
           citecolor=blue
           ]{hyperref}

\usepackage{kpfonts}
\DeclareMathAlphabet{\mathsf}{OT1}{cmss}{m}{n}
\SetMathAlphabet{\mathsf}{bold}{OT1}{cmss}{bx}{n}

\usepackage{fullpage}

\def\naiveabrev{naive-NN}
\def\naive{Naive Neural Network}
\def\encoderabrev{SE-NN}
\def\encoder{End-to-End Neural Network with Shared Encoder}

\begin{document}

\title{\huge Deep Learning Assisted End-to-End Synthesis of mm-Wave Passive Networks with 3D EM Structures: A Study on A Transformer-Based Matching Network\thanks{keywords: direct synthesis, deep learning, millimeter wave, impedance matching, transformer}}

\author{
Siawpeng Er{\textsuperscript{2}},
Edward Liu{\textsuperscript{1}},
Minshuo Chen{\textsuperscript{2}}, 
Yan Li{\textsuperscript{2}},\\\vspace{0.05in}
Yuqi Liu{\textsuperscript{1}}, 
Tuo Zhao{\textsuperscript{2}}, 
Hua Wang{\textsuperscript{1}}
\vspace{0.07in}
\\
{\textsuperscript{1}}School of Electrical and Computer Engineering, Georgia Tech\\
{\textsuperscript{2}}School of Industrial and Systems Engineering, Georgia Tech\\
}

\date{}
\maketitle

\begin{abstract}
This paper presents a deep learning assisted synthesis approach for direct end-to-end generation of RF/mm-wave passive matching network with 3D EM structures. Different from prior approaches that synthesize EM structures from target circuit component values and target topologies, our proposed approach achieves the direct synthesis of the passive network given the network topology from desired performance values as input. We showcase the proposed synthesis Neural Network (NN) model on an on-chip 1:1 transformer-based impedance matching network. By leveraging parameter sharing, the synthesis NN model successfully extracts relevant features from the input impedance and load capacitors, and predict the transformer 3D EM geometry in a 45nm SOI process that will match the standard 50$\Omega$ load to the target input impedance while absorbing the two loading capacitors. As a proof-of-concept, several example transformer geometries were synthesized, and verified in Ansys HFSS to provide the desired input impedance.
\end{abstract}


\section{Introduction}
Passive matching networks are used extensively in RF/mm-Wave circuits and  systems to ensure connections between devices/circuits with desired performance properties such as maximum power transfer, loadline matching, noise matching, and passive voltage/current scaling  \cite{razavi2012rf,pareview}.
Although there are many passive network topologies to perform impedance matching (e.g., \cite{wang2010cmos,hu2015design,mannem2020reconfigurable,nguyen2020coupler,liu2015calibrated,chi2018millimeter,huang202024,wang2019super,hu201928}),
one commonality is that the process of designing, verifying, and optimizing these networks requires extensive use of EM simulations, such as Ansys HFSS. As a result, the design of these networks is oftentimes very tedious, time-consuming, and requiring extensive design experience and computation resources. Only after constructing and simulating the EM structure can the designer then extract circuit performance metrics such as S-parameters, Z-parameters, loss, bandwidth, and examine the whole passive network performance. As the complexity of the passive network structure grows, simulation times may easily reach in the excess of several hours. Furthermore, this design and optimization process usually requires extensive iterations, further increasing the simulation time needed. Therefore, it will be extremely beneficial for a designer to have an end-to-end ``synthesis tool'' in which, given the passive network topology and the target passive network's performance metrics as the input, outputs the EM structure geometry.

\begin{figure}[t]
	\centering
	\includegraphics[width=0.8\textwidth]{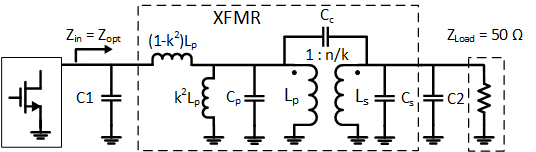}
	\caption{Schematic of transformer-based impedance matching network.}
	\label{fig:schematic}
\end{figure}

In this paper, we present an end-to-end neural network based method of direct synthesizing transformer-based impedance matching networks at $30$GHz by only providing circuit performance values, i.e., the desired impedance for matching and the loading capacitors $C_{1}$ and $C_{2}$. A schematic of the transformer-based impedance matching network considering the non-ideal magnetic coupling for power amplifier load impedance matching is shown in Fig. \ref{fig:schematic}. The capacitors $C_{1}$ and $C_{2}$ are used to transform the fixed load $R_{L}$ to some impedance $Z_{\rm in}$, and they are often specified by the load capacitors and device output capacitors that should be absorbed in the network.

\begin{figure}[t]
	\centering
	\includegraphics[width=0.8\textwidth]{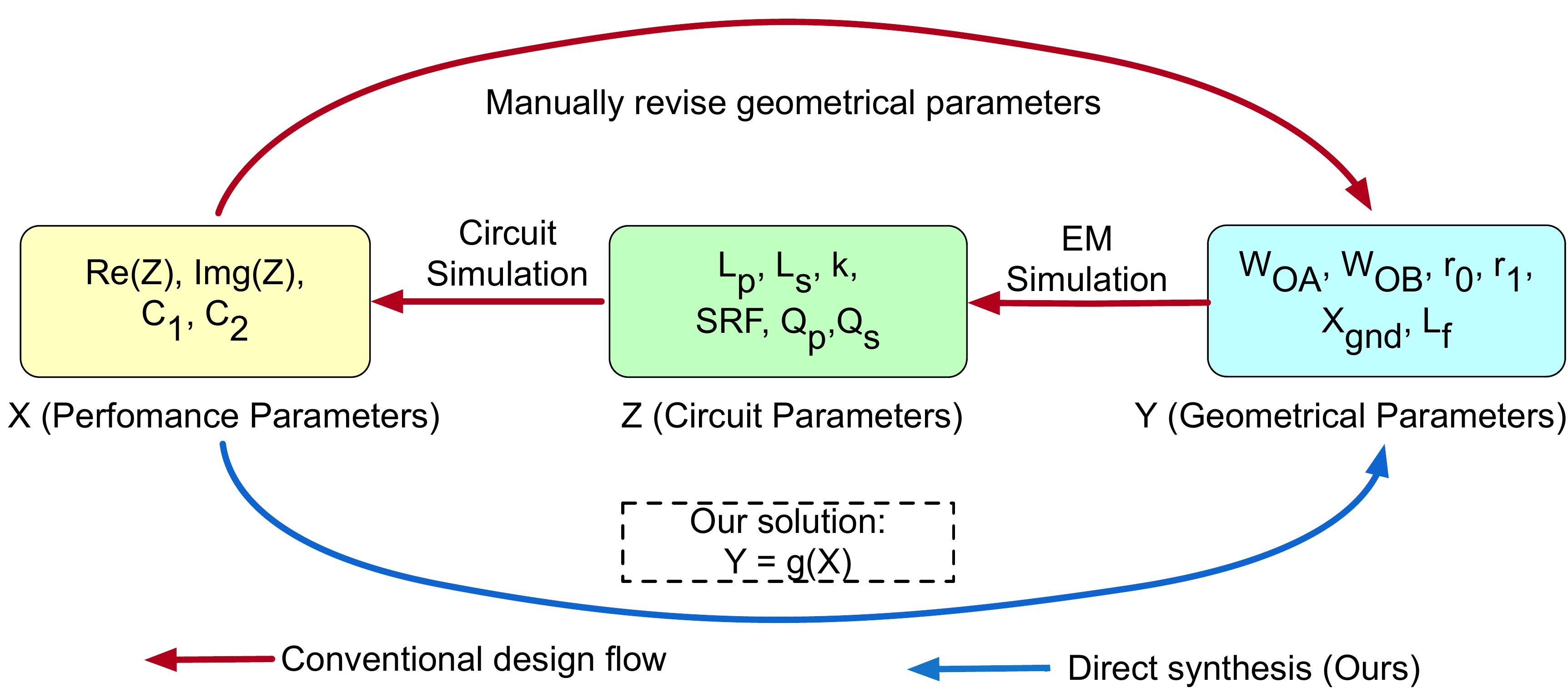}
	\caption{Comparison of existing iterative design cycle and our predictive model which directly outputs target geometry based on performance parameters.}
	\label{fig:designflow}
\end{figure}

By synthesizing the matching network including the EM structure directly from the required network performance parameters, e.g., the target load impedance, designers' efforts and computation resources required to iterate through EM geometries and network designs can be drastically reduced. Fig. \ref{fig:designflow} shows the comparison of the traditional design flow and our neural network-based approach. Given the desired circuit performance values --- input impedance $Z_{\rm opt}$, together with load capacitors $C_{1}$ and $C_{2}$ --- the synthesis network generates the desired transformer design parameters, including the coil radii ($r_0$ and $r_1$), widths ($W_{\rm OA}$ and $W_{\rm OB}$), ground spacing ($x_{\rm gnd}$), and input/output feed length ($\ell_{\rm f}$), achieving an end-to-end synthesis of transformer-based impedance matching network. Fig. \ref{fig:inout_model} shows the synthesized transformer structure, input parameters, and output physical parameters.

Our approach is closely related to the prior work in \cite{9218278}, where neural networks are used to synthesize $1$:$1$ on-chip transformers. However, \cite{9218278} only take the transformer circuit component values, not performance values, to predict the EM geometry, which is not an end-to-end synthesis solution.

\section{End-to-End Passive Network Synthesis Model}
We propose to use a Neural Network (NN) model to directly predict the geometry of the EM structure in the impedance matching network. In the recent surge of studies, overparameterized neural networks (the number of free weights exceeds that of training samples) are shown to exhibit extraordinary predictive power. Meanwhile, empirical results also suggest that large neural networks can be efficiently trained using modern algorithms (e.g., stochastic gradient descent), and surprisingly provide faithful fit to unseen data.

Although the outstanding performance of NNs is pronounced in various applications, the architectures of NNs need to be carefully designed to adapt to different tasks. 
In this end-to-end synthesis task, a major caveat is that the information degeneracy of predicting from a smaller input dimension ($4$ performance parameters) to a larger output dimension ($6$ physical parameters). In fact, as illustrated in Fig. \ref{fig:designflow}, once the physical parameters are given, the performance parameters are governed by the physical law, and can be viewed as a function of the physical parameters. By the data processing inequality, the information contained in the performance parameters may be less than that in the physical parameters, which incurs difficulty in the direct synthesis of the physical parameters.

To tackle such a difficulty, we propose an NN architecture with a shared encoder in Fig. \ref{fig:network}. We simultaneously predict the physical and circuit component parameters. In order to accurately predict the circuit parameters ({\textcolor{blue}{blue} part in Fig. \ref{fig:network}), the features extracted by the shared encoder need to replicate the relevant information in the circuit components.
	As a result, when using the same extracted features to predict the physical parameters, the shared encoder guides the interaction of the performance parameters and the physical parameters based on the underlying circuit and the physical law.

\section{Passive Matching Network Direct Synthesis}
\begin{figure}[t]
	\centering
	\includegraphics[width=0.8\textwidth]{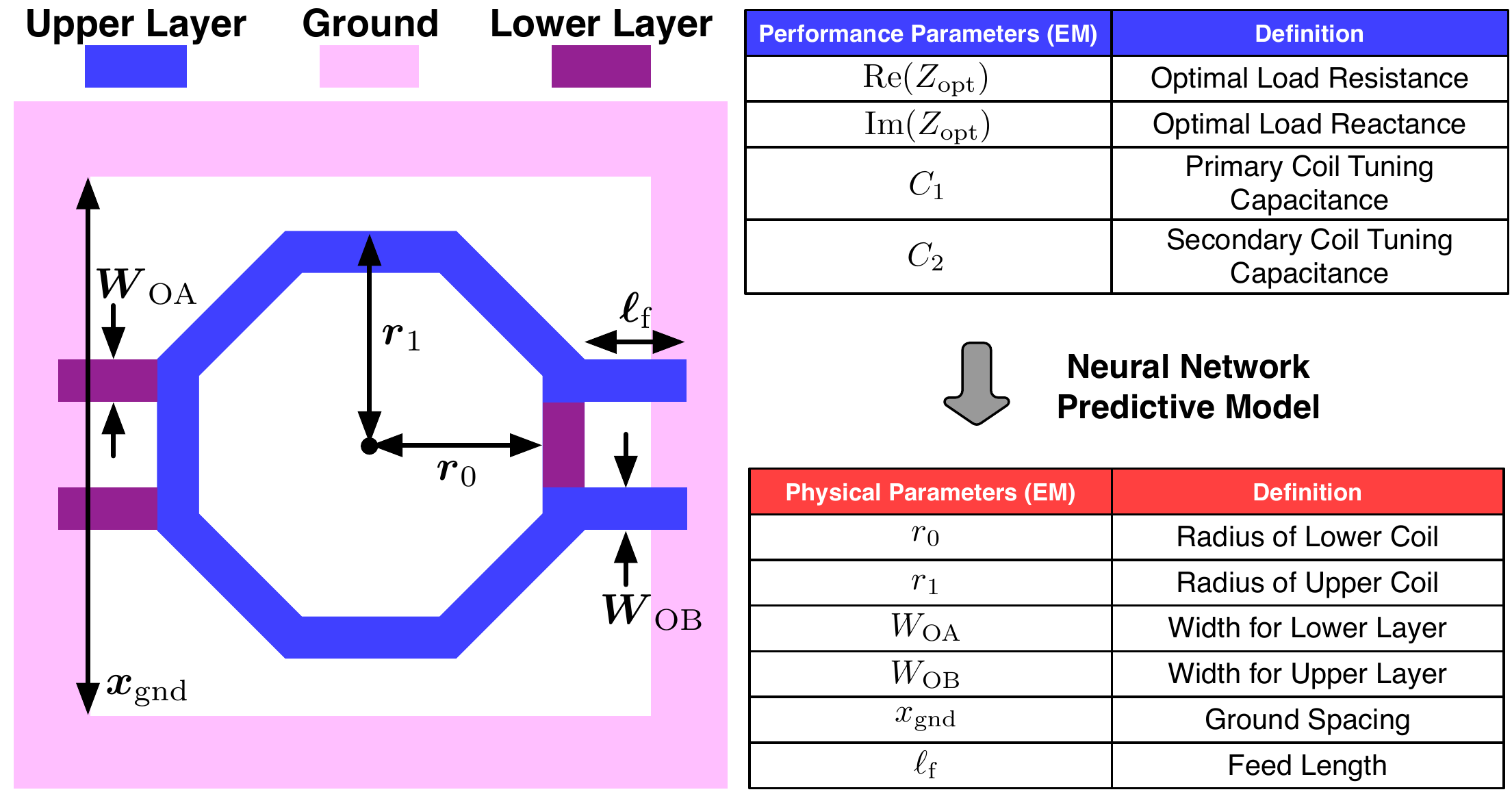}
	\caption{Input circuit performance metrics and output physical parameters with transformer EM model.}
	\label{fig:inout_model}
\end{figure}

\begin{figure}[t]
	\centering
	\includegraphics[width=0.8\textwidth]{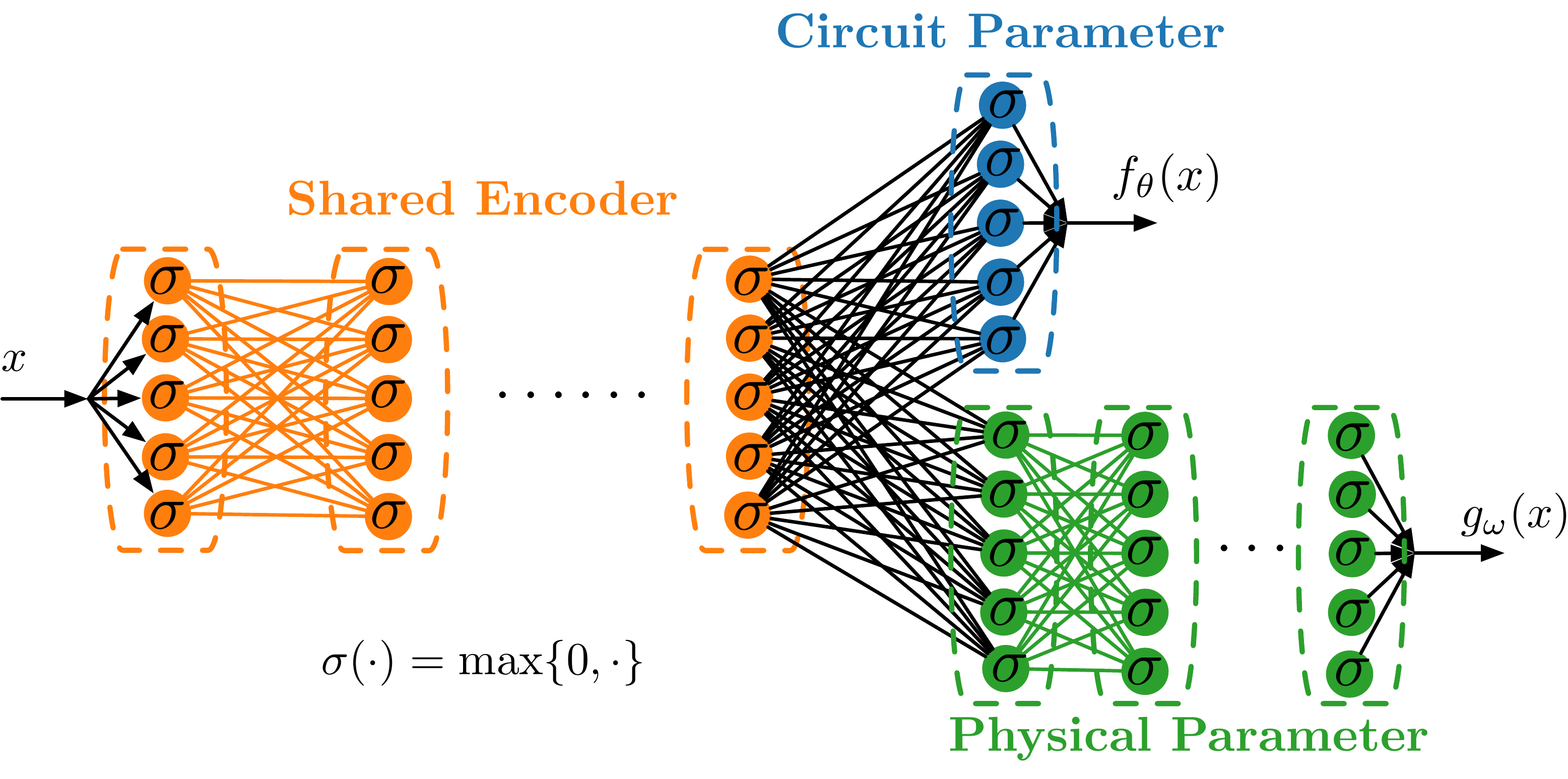}
	\caption{Synthesis model architecture. $f_\theta(x)$ predicts intermediate circuit parameters and $g_\omega(x)$ predicts the physical parameters.}
	\label{fig:network}
\end{figure}

\noindent $\bullet$ \textbf{Data Generation}.
We use Ansys HFSS simulation software to obtain $2.48$ million triples $(x_i, y_i, z_i)$, where $x = [{\rm Re}(Z_{\rm opt}), {\rm Im}(Z_{\rm opt}), C_1, C_2]^\top \in \mathbb{R}^4$ denotes the concatenation of the input impedance and load capacitors, $z \in \mathbb{R}^6$ denotes circuit parameters, and $y \in \mathbb{R}^6$ denotes physical parameters.
We shuffle all the triples and randomly select $1.99$ million as the training data, and the remaining $0.49$ million as the testing data.

\noindent $\bullet$ \textbf{Synthesis Network Architecture}.
As illustrated in Fig. \ref{fig:network}, our synthesis model (denoted as \encoderabrev~(\encoder)) consists of two sub-networks, both of which take $x$ as input. One sub-network $f_\theta$, with trainable weights $\theta$,  predicts circuit parameters $z$; another sub-network $g_\omega$, with trainable weights $\omega$,  predicts physical parameters $y$. 
The two sub-networks share parameters in the encoder for extracting features from the input.

Throughout our experiments, we choose $f_\theta$ as an $8$-layer feedforward NN and $g_\omega$ as a $11$-layer NN. The shared encoder is a $7$-layer feedforward NN. The activation function is the rectified linear unit, i.e., $\sigma(\cdot) = \max\{\cdot, 0\}$. Each layer within the synthesis network model contains $2048$ neurons.



\noindent $\bullet$ \textbf{Training Procedure}.
We introduce our loss functions and optimization algorithm for training the end-to-end synthesis model. Given $n$ targeted parameters $v_i \in \mathbb{R}^k$ (interpreted as either $z_i$ or $y_i$) and their corresponding predictions $\hat{v}_i$, we measure the average discrepancy using 1) Scaled Mean Squared Error (SMSE)
\begin{align*}
	\textstyle
	\textrm{SMSE}(\{\hat{v}_i, v_i\}_{i=1}^n) = \frac{1}{nk} \sqrt{\sum_{i = 1}^n \sum_{j = 1}^k \left(\frac{v_{i,j}  - \hat{v}_{i,j}}{ v_{i,j}} \right)^2};
\end{align*}
2) Scaled Dimensional Mean Squared Error (SDMSE) 
\begin{align}
	\textstyle
	\textrm{SDMSE}(\{\hat{v}_i, v_i\}_{i=1}^n) = \frac{1}{k} \sum_{j = 1}^k  \sqrt{\frac{1}{n}\sum_{i = 1}^n\left(\frac{v_{i,j}  - \hat{v}_{i,j}}{ v_{i,j}} \right)^2},
\end{align}
where $v_{i, j}$ denotes the $j$-th coordinate of $v_i$. We remark that both the evaluation metrics account for the different magnitudes of coordinates in $v$, while SDMSE promotes a balanced prediction performance across coordinates. 
We train our model by minimizing the following empirical risk:
\begin{align}
	\min_{\theta, \omega} L(\theta, \omega) = \Phi_n\left(\{g_\omega(x_i), y_i\}_{i=1}^n\right) + \lambda \Phi_n\left(\{f_\theta(x_i), z_i\}_{i=1}^n\right),
\end{align}
where $\Phi_n$ is the loss function, chosen as  either SMSE or SDMSE.
The hyper-parameter $\lambda$ controls the information to be retrieved in the shared encoder. We fine tune $\lambda$ and set to be $0.5$ across experiments. Note that when $\lambda=0$, the synthesis model will not benefit from the parameter sharing, since the error of $f_\theta$ is neglected.
From a statistical perspective, $\lambda \Phi_n\left(\{f_\theta(x_i), z_i\}_{i=1}^n\right)$ can be viewed as a regularizer, which encourages the synthesis model to encode the underlying physical law between between the performance, circuit, and physical parameters.

We use Adam to minimize the empirical risk $L(\theta, \omega)$. Adam is a modern stochastic first-order algorithm, known for its superior empirical performance in training neural networks. 
In each iteration of the algorithm, we randomly select a small number of training samples (mini-batch) to obtain a stochastic approximation of the gradient:
\begin{align*}
	\textstyle
	\widehat{\nabla} L(\theta, \omega) = \nabla \Phi_b\left(\{g_\omega(x_{i}), y_{i}\}_{i \in \mathcal{B}}\right) + \lambda \nabla \Phi_b\left(\{f_\theta(x_{i}), z_{i}\}_{i \in \mathcal{B}}\right) ,
\end{align*}
where $b \ll n$ is the size of the mini-batch $\mathcal{B}$ and $\nabla$ is understood as the gradient with respect to $\theta$ and $\omega$. The full algorithm is presented in Algorithm \ref{alg:adam}.

In our experiments, we set the mini-batch size $b = 1024$ and the initial learning rate $\eta = 0.001$. Every $50$ epochs, we decay the learning rate by half for a total of $500$ epochs. Other hyper-parameters, e.g., $\beta_1$ and $\beta_2$, are set as default in Adam.

\begin{algorithm}[h]
	\caption{Adam algorithm, $\sqrt{\cdot}$, $(\cdot)^{-1}$, and $\odot$ denote element-wise square root,  inverse,  and  multiplication.}
	\label{alg:adam}
	\begin{algorithmic}
		\STATE{\textbf{Input:} 
			learning rate $\eta$, $\beta_1, \beta_2 $},  $\epsilon $, weight decay $\tau$.
		\STATE{\textbf{Initialize:} $\theta_0$, $m_0 = 0$, $v_0 = 0$, $t = 0$.}
		\WHILE{$\theta_t$ not converged}
		\STATE{ Set $t = t+1$, choose mini-batch  $\mathcal{B} \subset \{1, \ldots, n\}$}
		\STATE{Calculate stochastic gradient $g_t = \widehat{\nabla} L(\theta, \omega)$ on $\mathcal{B}$.}
		\STATE{$m_t = \beta_1 m_{t-1} + (1-\beta_1) g_t$ }
		\STATE{$v_t = \beta_2 v_{t-1} + (1-\beta_2) g_t\odot g_t$}
		\STATE{$\hat{m}_t =  m_t / (1 - \beta_1^t)$, $\hat{v}_t = v_t / (1 - \beta_2^t) $}
		\STATE{$[\theta_t, \omega_t]^\top = (1 -\eta \tau) [\theta_{t-1}, \omega_{t-1}]^\top - \eta \hat{m}_t \odot (\sqrt{\hat{v}_t} + \epsilon )^{-1} $ }
		\ENDWHILE
		\RETURN{ $\theta_t$, $\omega_t$}
	\end{algorithmic}
\end{algorithm}

\noindent $\bullet$ \textbf{Predictive Performance}.
We compare the performance of our \encoderabrev~synthesis model with benchmark ML methods --- linear regression and gradient boosting. We also contrast with predicting the physical parameters using $g_\omega$ only without simultaneously training $f_\theta$ (i.e., $\lambda=0$). Such a model does not leverage the parameter sharing and is denoted as \naive\ (\naiveabrev) model. 

The performance is measured using SMSE and coefficient of determination ($R^2$). The latter is defined as
\begin{align}
	\textstyle
	R^2(\{\hat{v}_i, v_i\}_{i=1}^n) = 1 -\frac{\sum_{i=1}^n\sum_{j=1}^k(v_{i,j}  - \hat{v}_{i,j})^2}{\sum_{i=1}^n\sum_{j=1}^k(v_{i,j}  - \bar{v}_{\cdot,j})^2},
\end{align}
where $\bar{v}_{\cdot, j}$ is the mean of the $j$-th coordinate of $v$, and $v, \hat{v}$ are taken as the physical parameters $y$ and its prediction, respectively.
We demonstrate the appealing performance of \encoderabrev\ synthesis model  
in Table \ref{tab:experiments}. As can be seen, while the \naiveabrev\  approach overperforms benchmark ML methods, its performance pales in comparison to our \encoderabrev\ synthesis model.
This observation advocates our careful design of the knowledge-sharing encoder, from which we obtain a drastic gain in predictive accuracy. 

\begin{table}[H]
	\caption{Performance comparison of predicting physical parameters using circuit performance parameters.}
	\label{tab:experiments}
	\begin{center}
		\begin{tabular}{ | c || c | c || c | c |}
			\hline
			\multirow{2}{*}{Model}  & \multicolumn{2}{c ||}{SMSE Training Loss} & \multicolumn{2}{c|}{SDMSE Training Loss} \\ \addlinespace[-0.05em] \hhline{~----}
			&&&& \\[-1em]
			& SMSE & $R^2$ & SMSE & $R^2$ \\
			\hline
			Gradient Boosting  &  0.2250  & 0.2333   & - &- \\
			\hline
			&&&& \\[-1em]
			Linear Regression & 0.2984  & 0.1166 &-  & - \\
			\hline
			&&&& \\[-1em]
			\naiveabrev  & 0.0349 & 0.4697 &  0.0380  & 0.4562  \\
			\hline
			&&&& \\[-1em]
			\encoderabrev & 0.0047 & 0.8979 &  0.0054  & 0.9029  \\
			\hline
		\end{tabular}
	\end{center}
\end{table}

\noindent $\bullet$ \textbf{Validation Examples of Direct Synthesis}.
We deploy our trained \encoderabrev~model to predict the transformer geometry in the impedance matching network in Fig. \ref{fig:schematic}.
The target performance parameters and load capacitors' value are fed into the synthesis model (note that we only need access to $g_\omega$ in the deployment).
The predicted physical parameters of $4$ different instances are shown in Table \ref{tab:predict_experiments}. We verify the predicted transformer geometries by EM simulations to reproduce circuit parameters used to derive the actual performance parameters. 
We observe that the synthesized performance parameters provide a good estimation of the target transformer geometry.

\begin{table}[H]
	\caption{Direct synthesized examples of transformer geometry in the impedance matching network using \encoderabrev model.}
	\label{tab:predict_experiments}
	\vspace{-0.1in}
	\begin{center}
		\begin{tabular}{|c | c || c | @{ }c@{ } |}
			\hline
			\multicolumn{2}{|c ||}{Performance Parameters} & ${\rm Re}(Z_\textrm{opt})$(Ohm) & ${\rm Im}(Z_\textrm{img})$(Ohm)  \\ \hline
			\multirow{2}{*}{I} & Targeted & 32.110 & -3.995 
			\\ 
			& Synthesized & 30.125 & -6.444  \\\hline
			\multirow{2}{*}{II} & Targeted & 38.620 & -2.643 
			\\ 
			& Synthesized & 42.005 & -5.655   \\\hline
			\multirow{2}{*}{III} & Targeted & 24.209 & 8.792 \\
			& Synthesized & 26.347 & 10.645  \\\hline
			\multirow{2}{*}{IV} & Targeted & 42.794 &  -16.231 \\ 
			& Synthesized & 34.722 & -18.470  \\\hline
		\end{tabular}
	\end{center}
	\vspace{-0.1in}
	\begin{center}
		\begin{tabular}{|@{ }c@{ } || @{ }c@{ } | @{ }c@{ } |}
			\hline
			\multirow{3}{*}{Synthesized Geometry I}& $W_{\textrm{OA}} = 15.10 \mu$m & $r_0 = 41.00\mu$m \\
			& $W_{\textrm{OB}} = 11.73\mu$m & $r_1 = 47.99\mu$m \\
			& $x_{\textrm{gnd}} = 67.53\mu$m & $\ell_{\textrm{f}} = 14.70\mu$m \\\hline
			\multirow{3}{*}{Synthesized Geometry II}& $W_{\textrm{OA}} = 10.09 \mu$m & $r_0 = 42.97\mu$m \\
			& $W_{\textrm{OB}} = 13.13\mu$m & $r_1 = 44.97\mu$m \\
			& $x_{\textrm{gnd}} = 57.84\mu$m & $\ell_{\textrm{f}} = 34.44\mu$m \\\hline
			\multirow{3}{*}{Synthesized Geometry III}& $W_{\textrm{OA}} = 14.86 \mu$m & $r_0 = 44.79\mu$m \\
			& $W_{\textrm{OB}} = 11.98\mu$m & $r_1 = 51.61\mu$m \\
			& $x_{\textrm{gnd}} = 58.32\mu$m & $\ell_{\textrm{f}} = 24.49\mu$m \\\hline
			\multirow{3}{*}{Synthesized Geometry IV}& $W_{\textrm{OA}} = 15.06 \mu$m & $r_0 = 48.46\mu$m \\
			& $W_{\textrm{OB}} = 48.46\mu$m & $r_1 = 58.83\mu$m \\
			& $x_{\textrm{gnd}} = 59.94\mu$m & $\ell_{\textrm{f}} = 15.29\mu$m \\\hline
		\end{tabular}
	\end{center}
\end{table}

\section{Conclusion}
We propose a deep learning assisted model for the direct synthesis of a transformer-based \\impedance matching network, using only Z$_{\rm opt}$, $C_{1}$, and $C_{2}$ as input. For any passive network design, the problem statement begins with the geometry of EM structures and the ultimate goal is to meet specific performance target. Traditionally, it is very difficult to map directly between the two ends of the problem, geometry and performance, where designers have to go back and forth between EM geometries, circuit parameters and then evaluate the overall performance. Our proposed synthesis network achieves this end-to-end synthesis efficiently and effectively. Our methods can be potentially applied to synthesis of other passive networks, which is our on-going research.

\bibliographystyle{ieeetr}
\bibliography{references}

\end{document}